\pdfoutput=1
\documentclass[conference]{IEEEtran}
\IEEEoverridecommandlockouts

\usepackage{cite}
\usepackage{amsmath,amssymb,amsfonts}
\usepackage{algorithmic}
\usepackage{graphicx}
\usepackage{textcomp}
\usepackage[table]{xcolor}
\usepackage{fontawesome5}
\usepackage{ulem}
\usepackage{tikz}
\usepackage{url}
\usepackage{booktabs,multirow}
\usepackage{makecell}
\usepackage[hidelinks]{hyperref}
\usetikzlibrary{positioning, shapes.geometric, arrows.meta, backgrounds, calc, fit}

\definecolor{pBlue}{HTML}{1F4E79}
\definecolor{pTeal}{HTML}{1F6F6B}
\definecolor{pAmber}{HTML}{B45309}
\definecolor{pPlum}{HTML}{6B2D7A}
\definecolor{pGreen}{HTML}{2F6F3F}
\definecolor{pCoral}{HTML}{B91C1C}
\definecolor{pBgQuery}{HTML}{EEF3F9}
\definecolor{pBgSystem}{HTML}{F6F2EA}
\definecolor{pBgResp}{HTML}{EEF5EE}
\definecolor{pBgKB}{HTML}{F1ECF5}
\definecolor{pBgCard}{HTML}{FFFFFF}
\definecolor{pBorder}{HTML}{B5B5AD}
\definecolor{pBorderSoft}{HTML}{D7D7D0}
\definecolor{pText}{HTML}{1F1F1F}
\definecolor{pMuted}{HTML}{6B6B6B}
\definecolor{markColor}{HTML}{006B3C}
\definecolor{carlosColor}{HTML}{804000}

\def\BibTeX{{\rm B\kern-.05em{\sc i\kern-.025em b}\kern-.08em
    T\kern-.1667em\lower.7ex\hbox{E}\kern-.125emX}}
\begin{document}

\title{Enabling Privacy-Preserving Natural Language Access to Industrial Smart Meter Data}

\author{
\IEEEauthorblockN{Carlos Martín-Abellán}
\IEEEauthorblockA{\textit{Valencian Research Institute for Artificial Intelligence (VRAIN)} \\
\textit{Universitat Politècnica de València}\\
Valencia, Spain \\
cmarabe1@vrain.upv.es}
\and
\IEEEauthorblockN{Mark Kröll}
\IEEEauthorblockA{\textit{Know Center Research GmbH} \\
Graz, Austria \\
mkroell@know-center.at}
}

\maketitle

\begin{abstract}
Energy management officers and grid operators increasingly rely on smart meter data for anomaly detection, performance monitoring, and strategic decision-making. However, extracting actionable insights from large-scale, heterogeneous telemetry databases requires advanced SQL and data engineering expertise, creating a critical bottleneck for non-technical personnel. Furthermore, the strict data privacy regulations and security constraints inherent to critical energy infrastructure often prohibit the routing of sensitive consumer data through cloud-based Large Language Models (LLMs). To address these challenges, we present \textit{OpenMeter}, a schema-agnostic Multi-Agent System (MAS) designed to translate natural language queries into executable code and human-readable insights.  
This architecture lowers the technical barrier for day-to-day utility operations by empowering non-expert personnel to interact with complex databases through intuitive natural language dialogue, transforming raw telemetry into a highly valuable tool for routine grid workflows and strategic decisions.  
Multi-scenario ablation studies over this environment demonstrate that
full multi-agent orchestration shows its largest observed accuracy gains on
resource-constrained local hardware (+6.1\,pp EX-R over the strongest
single-agent configuration); at frontier scale, individual components
retain value but should be enabled selectively, indicating that the
cost of full orchestration is justified specifically where privacy
demands local execution.
\end{abstract}


\section{Introduction}

\noindent The rapid deployment of Advanced Metering Infrastructure has transformed modern power grids into highly data-intensive systems. While the continuous stream of telemetry holds immense potential for grid optimization and automated auditing, bridging the gap between raw database storage and actionable operational insights remains a significant challenge.

\textit{Industrial Motivation and the Data Access Problem:} 
Smart meters deployed at scale generate time-series data at
10--15\,minute intervals, covering active energy, voltage, reactive
power, and power quality metrics across hundreds of metering
points~\cite{wang2024}.
However, the volume and complexity of this telemetry creates what Dong et~al. describe as ``sensory overload'' for operational staff~\cite{dong2024}: while the data required for anomaly detection, standards compliance verification,
tariff auditing, and data availability assessment is successfully
captured, its analytical utility remains largely inaccessible to the
personnel who need it most.

Energy Management Officers are domain experts in grid operations, not database engineers,
and this mismatch creates a well-documented operational
bottleneck~\cite{quamar2022}: together with an industrial company providing smart energy solutions, we identified four recurring analytical tasks - i) reporting frequency checks, ii) data availability audits, iii) structural quality assessment, and iv) anomaly detection - which remain inaccessible to operations staff without database engineering support.

By enabling operators to communicate with their data in plain language, systems like \textit{OpenMeter} (see Fig.~\ref{fig:arch}) democratize access to grid analytics, reducing response latency from hours to minutes and supporting faster,
data-driven decision-making at the field level.

Compounding the access problem, the sensitive nature of consumer
energy data (from which adversaries can infer occupancy patterns,
daily routines, appliance usage, and even individual
identities~\cite{tonyali2018}, with user re-identification accuracy
reaching 99\% on public smart meter
datasets~\cite{zaman2025}) prohibits routing operational queries
through external cloud-based LLM APIs. Any practical solution for
this domain must therefore rely on open AI models deployed in
local, controlled environments within the utility's own
infrastructure, ensuring that sensitive consumption data never leaves
the operator's premises. %

To the best of our knowledge, no existing system integrates LLM-driven multi-agent orchestration, Text-to-SQL translation, local open-weight execution, and domain-grounded explainability for real smart meter data analytics. Our main contribution is as follows:

  \textbf{- A privacy-preserving multi-agent system for energy Text-to-SQL with domain-grounded explainability.} We propose, \textit{OpenMeter}, the first system combining multi-agent LLM orchestration with Text-to-SQL for smart meter data analytics. Co-designed with an industrial partner to address daily operational tasks, the system is grounded in three design principles: (a)~\textit{Privacy-by-Architecture}, (b)~\textit{Agent Separation} and (c)~\textit{LLM-as-Curator}.

  \noindent As supporting contributions, we provide: \\
  \noindent - An \textbf{additive ablation study}
  (\textsc{S-Base}, \textsc{S+SC}, \textsc{M-Base}, \textsc{M+SC},
  \textsc{M+KB}, \textsc{M+Full}) that independently isolates the
  contributions of self-correction, multi-agent decomposition, and
  domain-knowledge injection, providing the first controlled
  single-agent versus multi-agent comparison for energy data querying.

  \noindent - An \textbf{energy-domain Text-to-SQL evaluation dataset} of
  62~verified NL-to-SQL pairs derived from operational workflows
  identified with the industrial partner's field engineers. Each pair underwent a
  three-step verification protocol: (1)~manual Gold SQL authoring; (2)~execution verification with non-empty result set inspection and (3)~SHA-256 result hashing for deterministic automated comparison.

\section{Background and Related Work}

\subsection{Multi-Agent Systems and LLM Orchestration}

Multi-Agent Systems (MAS) have long been used for distributed energy management. 
Ali et al.~\cite{ali2025} propose a hierarchical MAS with specialized agents for resource scheduling, yet these systems operate exclusively as backend controllers and lack natural language interfaces, preventing human operators from conversing with or querying them directly.

The emergence of LLM-based multi-agent systems has begun to address this interaction gap through role specialization. Xiao and Xu~\cite{xiao2024} present the first LLM-driven MAS applied to building energy auditing, deploying three specialized agents (Planner, Researcher, Advisor) to automate the diagnosis of energy inefficiencies.
This ``Orchestrator-Programmer-Reviewer" pattern has been formalized in software engineering (cf. He et al.~\cite{he2025}) for automated code generation and validation. There, task decomposition, implementation, and quality assurance are structurally separated across agents, a paradigm that closely mirrors our 
system pipeline (see Fig.~\ref{fig:arch}).

In addition, existing energy-specific LLM frameworks, such as Gamage et al.'s multi-agent RAG system for net-zero management~\cite{gamage2024}, rely on cloud infrastructure and lack Text-to-SQL capabilities.

To our knowledge, no prior work combines MAS orchestration with Text-to-SQL for energy analytics while providing a controlled comparison between single-agent and multi-agent pipelines, which our ablation study addresses.

\subsection{Text2SQL for Structured Data Analytics}

The effectiveness of LLM-driven Text-to-SQL critically depends on how the
database schema is presented to the model. Gao et al.~\cite{gao2023} show that representing database schemas as SQL \texttt{CREATE TABLE} statements provides the most effective prompting strategy for Text-to-SQL models. 
Ojuri et al.~\cite{ojuri2025} further demonstrate the practical value of coupling intelligent agents with LLMs to democratize database access
for non-technical business users.

However, high performance on curated benchmarks does not transfer to real-world
deployment. Evaluations on Spider~2.0~\cite{lei2024} show that the best text-to-SQL parser achieves
only 5.7\% EX, while the strongest agent-based approach reaches just 21.3\%. 
The BIRD benchmark~\cite{li2024bird} further establishes that
resolving queries on noisy databases requires integrating external domain
knowledge beyond the schema itself, a finding that directly motivates our LLM-as-Curator mechanism.
Nascimento et al.~\cite{nascimento2025} corroborate this on IndDB, a proprietary industrial database: most SOTA prompting strategies fail to exceed 30\% EX, with the best achieving only 41\%. Crucially, their error analysis identifies schema linking as the dominant failure mode, accounting for 53.3\% of all errors, confirming that models routinely hallucinate table and column references even when the complete schema is provided in the prompt. 
Our system directly addresses this vulnerability through a deterministic schema validation layer backed by an iterative self-correction loop (Section~\ref{sec:arch}).

Within energy analytics, Intelli-Dispatch-SQL~\cite{ni2025} represents the closest related work, employing a single LLM agent for intent recognition, SQL generation, and validation to answer natural language
queries over power dispatching databases. Unlike OpenMeter, however, it operates on a synthetic dataset and a monolithic single-agent architecture without explainability metrics. \textit{OpenMeter} addresses these limitations through a multi-agent pipeline with structural separation between data querying and explanation generation,
evaluated on real industrial smart meter telemetry.

\subsection{LLMs in Energy Analytics and Privacy}

Data confidentiality remains one of the main barriers to deploying LLMs in energy systems. Dong et al.~\cite{dong2024} warn that cloud-hosted LLMs introduce significant cybersecurity and privacy risks. Similarly, Madani et al.~\cite{madani2025} identify data privacy as the main technical challenge and observe that the vast majority of proposed use cases remain conceptual, with no deployed systems operating on real grid infrastructure.

At the same time, explainability in smart meter analytics remains limited to post-hoc XAI techniques designed for traditional machine learning models~\cite{machlev2022}, providing little support for natural-language interaction.

Existing energy-focused LLM systems, such as Gamage et al.~\cite{gamage2024}, improve decision support through multi-agent architectures but continue to rely on cloud-hosted GPT-4, inheriting the privacy concerns identified above.

\textit{OpenMeter} addresses both challenges by executing all multi-agent inference locally with open-weight models while generating natural-language explanations grounded in domain rules rather than post-hoc feature attribution.

\section{Multi-Agent System Architecture}
\label{sec:arch}

\begin{figure*}[h!]
    \centering
    \resizebox{0.8\textwidth}{!}{%
    \begin{tikzpicture}[
      remember picture,
      >={Stealth[length=4pt,width=3pt]},
      iobox/.style    ={rectangle, rounded corners=3pt, draw=pBorder,
                        line width=0.45pt, inner sep=4pt, align=left,
                        text width=3.0cm, font=\sffamily\tiny},
        respbox/.style  ={rectangle, rounded corners=3pt, draw=pBorder,
                          line width=0.45pt, inner sep=4pt, align=left,
                          text width=3.2cm, font=\sffamily\tiny},
      agent/.style    ={rectangle, rounded corners=2pt, draw=pBorderSoft,
                        line width=0.35pt, fill=pBgCard, inner sep=3pt,
                        align=center, text width=2.4cm,
                        font=\sffamily\tiny},
      kbbox/.style    ={rectangle, rounded corners=2pt, draw=pBorderSoft,
                        line width=0.35pt, fill=pBgKB, inner sep=3pt,
                        align=center, text width=3.0cm,
                        font=\sffamily\tiny},
      dbbox/.style    ={cylinder, shape border rotate=90, draw=pBorderSoft,
                        line width=0.35pt, fill=pBgSystem, inner sep=3pt,
                        align=center, text width=1.5cm, minimum height=1.2cm,
                        font=\sffamily\tiny, aspect=0.25},
      mainflow/.style ={->, line width=1.0pt, draw=pBlue!85,
                        shorten >=2pt, shorten <=2pt},
      subflow/.style  ={->, line width=0.5pt, draw=pMuted,
                        shorten >=2pt, shorten <=2pt},
      dataflow/.style ={<->, line width=0.4pt, draw=pAmber!85, dashed,
                        shorten >=2pt, shorten <=2pt},
      lbl/.style      ={font=\sffamily\tiny\itshape, text=pText,
                        inner sep=1pt, align=center,
                        rounded corners=1pt},
    ]

    \node[iobox, fill=pBgQuery] (user) at (-0.8, 0)
    {{\color{pBlue}\faUserTie}~{\color{pBlue}\bfseries Operator}\\[2pt]
     \itshape Which meters recorded phase voltages below
     207\,V for more than 1\% of measurements in January 2025?};

    \node[agent] (orch) at (3.6, -0.75)
      {{\color{pTeal}\faSitemap}\;\textbf{Orchestrator (O)}\\[1pt]
       Intent\;\textbullet\;Plan\;\textbullet\;Context injection};

    \node[agent] (ca) at (8.4, -0.75)
      {{\color{pAmber}\faCode}\;\textbf{Coding Agent (CA)}\\[1pt]
       Text-to-SQL\;\textbullet\;Python};

    \node[dbbox] (db) at (8.4, -3.15)
      {{\color{pAmber}\faDatabase}\\[1pt]
       \textbf{DuckDB}\\[1pt]
       Python sandbox};

    \node[kbbox] (kb) at (3.6, 0.75)
      {{\color{pPlum}\faBookOpen}\;\textbf{Domain KB \& Schema}\\[1pt]
       Glossary\;\textbullet\;structured schema\;\textbullet\;rules};

    \node[agent] (ia) at (3.6, -3.15)
      {{\color{pPlum}\faLightbulb}\;\textbf{Interpretation Agent (IA)}\\[1pt]
       Synthesis\;\textbullet\;Explanation};

    \node[respbox, fill=pBgResp] (resp) at (-0.8, -2.4)
    {{\color{pGreen}\faClipboardCheck}~{\color{pGreen}\bfseries Grounded Response}\\[3pt]
     \centering
     \setlength{\tabcolsep}{4pt}
     \renewcommand{\arraystretch}{1.05}
     \begin{tabular}{@{}lrr@{}}
     \toprule
     \textbf{Meter} & \textbf{Viol.} & \textbf{Rate} \\
     \midrule
     ISKRA-0012 & 643 & \cellcolor{pCoral!22}\textbf{3.2\,\%} \\
     ISKRA-0045 & 412 & \cellcolor{pCoral!12}\textbf{2.1\,\%} \\
     \bottomrule
     \end{tabular}\\[4pt]
     \raggedright\itshape
     Both meters exceed the EN\,50160 undervoltage limit; ISKRA-0012 is
     the more severe case (3.2\,\%). A cabling fault or overloaded
     transformer is the likely cause; a site inspection is recommended.};

  \begin{scope}[on background layer]
  \coordinate (rightpad) at ([xshift=25pt]ca.east);
  \node[draw=pBorderSoft, rounded corners=4pt,
        line width=0.4pt, dashed,
        fill=pBgSystem,
        fit=(kb)(orch)(ca)(db)(ia)(rightpad),
        inner xsep=8pt, inner ysep=8pt] (sysbox) {};
  \end{scope}

  \node[anchor=south west, font=\sffamily\tiny]
    at ([xshift=2pt, yshift=2pt]sysbox.north west) {%
      {\color{pBlue}\faRobot}~{\color{pBlue}\bfseries OpenMeter}%
  };

    \draw[mainflow] (user.east) -- ++(0.6,0) |- (orch.west);
    \draw[subflow] (kb.south) -- (orch.north)
        node[lbl, midway, right=2pt] {Retrieval};

    \draw[mainflow]
        ([yshift=2pt]orch.east) -- ([yshift=2pt]ca.west)
        node[lbl, midway, above=1pt] {Plan + filtered rules};

    \draw[subflow]
        ([yshift=-2pt]ca.west) -- ([yshift=-2pt]orch.east)
        node[lbl, midway, below=1pt] {Result (DataFrame\,/\,fail)};

    \draw[dataflow]
        ([xshift=-3pt]ca.south) -- ([xshift=-3pt]db.north)
        node[lbl, midway, left=2pt] {SQL\,/\,Python};

    \draw[subflow]
        ([xshift=3pt]db.north) -- ([xshift=3pt]ca.south)
        node[lbl, midway, right=2pt] {Data\,/\,trace};

    \draw[->, line width=0.4pt, draw=pAmber!85, dashed,
          shorten >=2pt, shorten <=2pt]
        ([xshift=-5pt]ca.north east)
        arc[start angle=0, end angle=180,
            x radius=0.4cm, y radius=0.4cm]
        node[lbl, pos=0.5, above=1pt]
          {Self-correction\\(schema validation)};

    \draw[mainflow]
        (orch.south) -- (ia.north)
        node[lbl, midway, right=2pt]
          {Result-aware\\context + raw};

    \draw[mainflow] (ia.west) -- ++(-0.6,0) |- (resp.east);

    \end{tikzpicture}%
    }%
    \caption{\textbf{OpenMeter multi-agent architecture and running example.}
    A natural-language query flows through the $\mathsf{O}\!\rightarrow\!\mathsf{CA}\!\rightarrow\!\mathsf{O}\!\rightarrow\!\mathsf{IA}$
    pipeline, producing a grounded response with retrieved results and
    domain-based interpretation.
    Blue: primary execution flow; grey: retrieval and feedback;
    dashed amber: DuckDB interaction and CA self-correction;
    dashed boundary: controlled runtime environment.}
    \label{fig:arch}
    \end{figure*}
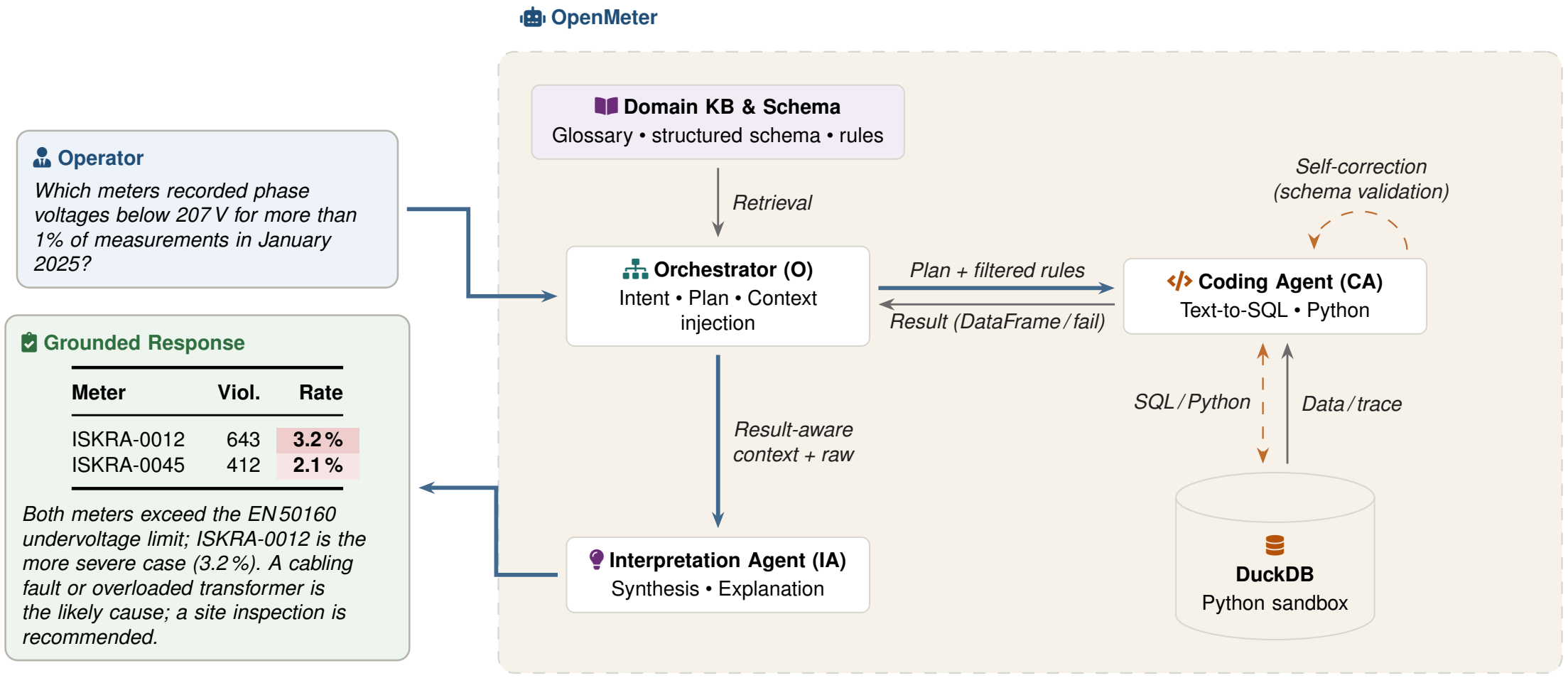

\subsection{System Overview and Design Principles}

Our Multi-Agent System (MAS) processes natural-language queries over smart
meter data and returns analytical results with domain-grounded explanations.
The architecture follows a \textit{Plan-then-Execute}
pattern~\cite{feng2024} (the Orchestrator produces a structured execution
plan before any code generation, preventing the premature-execution failure
mode reported for autonomous agents~\cite{cemri2025,he2025}) and is built
on three design principles:

\noindent\textit{(i) Privacy-by-Architecture:} All LLM inference runs on
open-weight models, either on-premises or via a privacy-preserving cloud
tier with prior data anonymization. Telemetry, queries, and results never
leave the controlled environment, satisfying the data-sovereignty
requirements of critical energy infrastructure.

\noindent\textit{(ii) Agent Separation:} The Coding Agent (CA), responsible
for data extraction via SQL or Python, and the Interpretation Agent (IA),
dedicated to natural-language generation, are structurally decoupled so that
raw numerical outputs are always mediated by a dedicated interpretation step.
This mitigates the Reasoning-Action Mismatch failure mode that
Cemri et al.~\cite{cemri2025} link to 13.2\% of execution errors in
LLM-based MAS benchmarks.

\noindent\textit{(iii) LLM-as-Curator:} The Orchestrator ($O$) manages the
domain knowledge base and routes only the relevant context per query,
removing the need for external RAG infrastructure (embedding models, vector
databases) and optimizing context-window utilization.

The execution flow is a sequential $O \to \mathrm{CA} \to O \to \mathrm{IA}$
pipeline (Fig.~\ref{fig:arch}), detailed agent by agent in the following
subsections.\\

\noindent \textit{Orchestrator Agent ($O$)} is the central coordinator of the MAS. It ingests the
natural-language query, the structural schema dictionary, and two
dataset-agnostic domain repositories grounded in established standards (EN~50160~\cite{en50160}, IEC~62056-21/DLMS~\cite{iec62056}): a
domain glossary of $\sim$45 smart-metering entries across 7~semantic
categories (each with term, definition, unit, and anomaly indicators), and a
domain knowledge base (KB) organized in 12~thematic categories that
formalizes operational expectations, cross-variable correlations, and
machine-checkable anomaly rules with severity levels, validated by professional
energy domain experts.

Before code generation ($O \to$ CA), $O$ performs four sequential operations:
intent classification into one of 13 analytical categories, difficulty
assessment (Easy/Medium/Hard), generation of a structured execution plan
(target channels, thresholds, visualization requirements), and selection of
the glossary entries and KB sections relevant to the classified intent. After
execution (CA $\to O \to$ IA), $O$ runs a second filtering pass over CA's raw
output and forwards to the Interpretation Agent (IA) only the domain rules
needed to contextualize the result.

\noindent\textit{LLM-as-Curator.} Injecting the full knowledge base into
every downstream prompt induces \textit{attention
degradation}~\cite{huang2025,nascimento2025}, where the model fails to
isolate the operative constraints from unrelated context. Rather than
truncating the KB or adding a retrieval pipeline, $O$ acts as a dynamic
per-query filter: it receives the full KB in its system prompt and
extracts only the applicable sections within its regular inference pass,
at no added latency and without the embedding model or vector store of a
RAG pipeline. Unlike Ni et al.'s~\cite{ni2025} fine-tuned lexicons, the
KB remains a version-controlled document set, updatable without
retraining. The contribution of this strategy is isolated in
\textsc{M-Base} vs.\ \textsc{M+KB} (Section~\ref{sec:setup}).\\

\noindent \textit{Coding Agent (CA)} is the Text-to-SQL generation engine. It receives from
$O$ the execution plan, the filtered domain rules, and the dataset schema.
Although Gao et~al.~\cite{gao2023} identify Code Representation
(\texttt{CREATE TABLE} syntax) as optimal for normalized databases, it is
inefficient for heterogeneous industrial deployments with EAV and wide-matrix
layouts. The CA therefore receives the schema as a \textit{structured context
block} with column names, types, cardinality statistics, enumerated value
domains, and sample rows, maximizing information density across multiple and diverse schemas.

The CA emits DuckDB-compatible SQL; for queries requiring graphical output, a
secondary Python stage produces matplotlib visualizations after successful SQL
execution. All analytical computation runs in SQL; Python is reserved for
visualization. Code executes in a sandboxed, read-only environment with no
network or filesystem access. The CA also implements an autonomous
\textit{self-correction loop} fully abstracted from $O$: on execution error,
DuckDB's column-level diagnostics are appended verbatim to the CA's context
for targeted refinement, directly addressing the schema-linking failure mode
reported in~\cite{nascimento2025}. $O$ only sees the final deterministic
outcome, preventing context pollution and cascading errors.\\

\noindent \textit{Interpretation Agent (IA)} is the final synthesis layer. It ingests the
original query, the raw output returned by CA, and the result-aware domain
context curated by $O$ post-execution. The IA produces a structured response
combining: a factual summary anchored to the retrieved values;
anomaly characterization that distinguishes statistical
anomalies (values outside expected ranges) from logical ones
(physically or operationally impossible values, e.g.\ negative cumulative
energy), citing the matching KB rule (e.g.\ V-001, E-005) when applicable;
and uncertainty-aware language, using hedged language for causal
inferences (``may indicate'', ``is consistent with'') and deterministic
language for retrieved facts. Visualization is triggered by the operator or inferred by $O$ at query time, captured in the execution plan, and rendered by the CA; the IA's role is exclusively explanatory.

\section{Experimental Setup}
\label{sec:setup}

\subsection{Six-Scenario Ablation Design}
\label{subsec:ablation}

The evaluation comprises six scenarios organized into two tracks: a single-agent baseline (S-Base, S+SC) and a multi-agent track (M-Base, M+SC, M+KB, M+Full). The single-agent track establishes the performance limits of a monolithic architecture, whereas the multi-agent track isolates the contribution of self-correction and domain knowledge through an additive $2\times2$ factorial design. %

\subsubsection{Single-Agent Track}
A single-agent configuration executes the full NL$\rightarrow$SQL$\rightarrow$interpretation pipeline within a unified prompt using the complete schema and domain knowledge.

\begin{itemize}
  \item \textit{S-Base:} A monolithic agent receives the full schema and domain KB without self-correction. This serves as the baseline for assessing end-to-end performance without architectural decomposition.

  \item \textit{S+SC:} Extends S-Base by enabling a bounded self-correction loop (maximum two retries upon execution failure), isolating the effect of iterative error recovery in a single-agent setting.
\end{itemize}

\subsubsection{Multi-Agent Track}

All multi-agent configurations employ the Orchestrator, Coding Agent, and Interpretation Agent. Starting from M-Base, self-correction and domain knowledge are independently enabled following a $2\times2$ factorial design.

\begin{itemize}
  \item \textit{M-Base (Baseline):} No self-correction and no domain knowledge injection.
  
  \item \textit{M+SC:} Enables the self-correction loop while keeping domain knowledge disabled.
  
  \item \textit{M+KB:} Enables selective domain knowledge routing while disabling self-correction.
  
  \item \textit{M+Full:} Combines domain knowledge injection and self-correction, representing the fully enabled multi-agent configuration.
\end{itemize}

The experimental design allows us to isolate the marginal contribution of self-correction and domain knowledge, as well as their potential interaction effects under a controlled factorial setup.

\subsubsection{Model and Hardware Configuration}

The Qwen~3.5 family (released February 2026) was selected as
it represents one of the highest-performing open-weight model
families available at the time of evaluation, and provides a
coherent parameter-scaling ladder from 9B to 397B-A17B~MoE
within a unified inference backend (Ollama), enabling controlled
scaling comparisons under identical infrastructure. All
experiments are conducted under two hardware tiers
(Table~\ref{tab:hardware}).

\begin{table}[t]
\centering
\caption{Hardware and model configuration for the evaluation tiers.}
\label{tab:hardware}
\renewcommand{\arraystretch}{1.2}
\setlength{\tabcolsep}{4pt}
\scriptsize
\begin{tabular}{@{}lcc@{}}
\toprule
 & \textbf{Setup~1 (Local)} & \textbf{Setup~2 (Cloud)} \\
\midrule
Model (all agents) & Qwen3.5 & Qwen3.5 \\
Parameters & 9B & 397B-A17B MoE \\
GPU & 1$\times$ RTX A5000 (24\,GB) & Ollama Cloud\textsuperscript{$\dagger$} \\
System RAM & 25\,GB & --- \\
vCPUs & 9 & --- \\
Inference backend & Ollama (local) & Ollama Cloud \\
Model swap overhead & None (unified model) & None (unified model) \\
\bottomrule
\end{tabular}
\vspace{2pt}
\par\noindent\raggedright\scriptsize
\textsuperscript{$\dagger$}\,Cloud hardware specifications are not
disclosed by the provider.
\vspace{-10pt}
\end{table}

\subsection{Datasets and Data Partitioning}
\label{subsec:datasets}
All experiments reported in Section~\ref{sec:results_and_evaluation} use an industrial smart-meter deployment operated by an industrial partner\footnote{The evaluation gold-standard dataset (62 NL-to-SQL pairs) and the measurement tables are available on Zenodo upon request: \url{https://doi.org/10.5281/zenodo.23983322}.}. The evaluation corpus comprises three measurement tables (EMSify, LP3 and LPR) totaling approximately 7.5\,million rows across EAV and Wide schemas (Table~\ref{tab:datasets}). For cloud-tier experiments, all personally identifiable information was anonymized before transmission, whereas local-tier experiments were executed entirely on-premises.

\begin{table}[b]
\vspace{-4pt}
\centering
\caption{Evaluation datasets (the industrial partner's deployment).}
\label{tab:datasets}
\renewcommand{\arraystretch}{1.2}
\setlength{\tabcolsep}{3pt} %
\scriptsize %
\begin{tabular}{@{}l c c c l p{2.8cm}@{}}
\toprule
\textbf{Dataset} & \textbf{Meters} & \textbf{Rows} & \textbf{Res.} & \textbf{Schema} & \textbf{Key Measurements} \\
\midrule
EMSify & 19 & ${\sim}$3.5M & 15m & Long/EAV & $A^{+}\!/A^{-}$, $R^{+}\!/R^{-}$, PF,
THD, $U_{L1\text{--}L3}$ (28 types) \\
LP3    & 74 & ${\sim}$3.0M & 10m & Wide     & $U_{L1\text{--}L3}$ (voltage only) \\
LPR    & 39 & ${\sim}$1.0M & 15m & Wide & $A^{+}\!/A^{-}$, load profile registers \\
\bottomrule
\end{tabular}

\vspace{4pt}
\begin{minipage}{\linewidth}
    \scriptsize
    Datasets span periods between 2019 and 2026 and exhibit distinct schemas and meter populations.
\end{minipage}
\end{table}

\subsubsection{Industrial Evaluation Dataset}
The industrial dataset consists of three measurement tables ingested without cleaning or imputation, preserving real telemetry artifacts (e.g., communication outages, counter resets, and undervoltage events). This enables evaluation under realistic deployment conditions rarely represented in academic benchmarks.

\subsubsection{Evaluation Golden Dataset (62 pairs)}
The golden dataset was derived from 39 natural-language templates covering 13 analytical categories and three difficulty levels. Each template was instantiated only for compatible schemas, yielding 62 manually verified NL$\rightarrow$SQL pairs using anonymized identifiers of real meters, and time ranges. Of these, 13 pairs were reserved for prompt refinement and debugging, while the remaining 49 were held out for the evaluation reported in Section~\ref{sec:results_and_evaluation}.

\subsection{Evaluation Metrics}
\label{subsec:metrics}
Metrics are organized into four categories: task accuracy, qualitative
assessment, self-correction efficiency, and pipeline performance. Our task accuracy metrics build on the standard Text-to-SQL benchmarking
protocol~\cite{gao2023,li2024bird}, which we extend with graded execution variants and a semantic-equivalence adjustment.

\subsubsection{Task Accuracy}\mbox{}\par\noindent

\textit{Runtime viability.} The \textit{Executable SQL Rate} (\textbf{ESR})
is the fraction of generated SQL queries that execute without syntax or database engine errors, whereas the \textit{Visualization Execution Rate} (\textbf{VER}) is the analogous rate for \texttt{matplotlib} scripts producing a valid image object. ESR serves as the prerequisite for all downstream evaluations.

\textit{Execution accuracy.} Generated and gold-standard result sets are
compared under three progressively relaxed variants that separate semantic
correctness from formatting differences. \textbf{EX-Strict (EX-S)}
requires identical column count, ordering, and row content (float
tolerance $\pm0.01$). \textbf{EX-Partial (EX-P)} relaxes column ordering
while preserving exact column and row counts. \textbf{EX-Relaxed (EX-R)},
our primary metric, performs name-blind value-based subset matching:
aliases are ignored, auxiliary columns are allowed, and all column
subsets and permutations are considered, provided the row count matches.
EX-R reflects operational settings where correct queries legitimately
return additional context (e.g., grouping keys) absent from the minimal
gold query. The differences
$\Delta(\mathrm{EX\text{-}P} - \mathrm{EX\text{-}S})$ and
$\Delta(\mathrm{EX\text{-}R} - \mathrm{EX\text{-}P})$
capture column-ordering errors and the effect of semantically valid
auxiliary information, respectively.

\textit{Structural equivalence.} \textit{Exact Match} (\textbf{EM})
compares normalized SQL abstract syntax trees (ASTs): both queries are
parsed using the DuckDB dialect, identifiers are normalized, aliases
removed, and order-insensitive clauses lexicographically sorted before
comparison. \textit{Component Matching} (\textbf{CM}) reports the mean
token-level $F_1$ after alias removal across the major SQL clauses,
yielding a continuous score in $[0,1]$.

\textit{Semantic Equivalence Rate (\textbf{SER}).} SER accounts for valid alternative SQL formulations that differ from the reference query but produce an equally acceptable interpretation. Candidate queries (ESR=1, EX-R=0) are filtered using majority consistency across runs before being reviewed by either an LLM judge or human evaluators. EX-S therefore serves as a conservative lower bound on true accuracy;
the SER acceptance rates reported in Table~\ref{tab:ser-results}
indicate how much of the remaining gap is attributable to valid
alternative formulations rather than genuine errors.

\subsubsection{Qualitative Assessment}\mbox{}\par\noindent

All qualitative dimensions are scored on a $[1,5]$ Likert scale.
\textit{Interpretation quality} is assessed along three axes:
\textbf{Groundedness (Grd)}, the factual traceability of every claim in
the response to the SQL result set;
\textbf{Relevance (Rel)}, the degree to which the response directly and
completely answers the target question without redundant diversions; and
\textbf{Coherence (Coh)}, internal consistency and logical flow. \textit{Visualization quality} is scored on
\textbf{Data Alignment (Align)}, the faithfulness of the chart to the
underlying SQL result set (correct axes, ranges, and visual encoding), and
\textbf{Presentation Quality (Pres)}, the legibility of labels, titles,
and legends for operator inspection.

\subsubsection{Self-Correction Efficiency}\mbox{}\par\noindent

The \textit{Self-Correction ROI} measures the net benefit of the
self-correction loop along two axes: $\Delta$ESR, the percentage-point
gain in executable queries after retries, and $\Delta$EX-R, the gain in
semantically correct queries. By design, the loop retries only after
execution errors, so regression of initially correct queries is
structurally impossible.

\subsubsection{Pipeline Performance}

\textit{End-to-end latency} $t_{\text{e2e}}$ is the wall-clock
duration from query input to final explanation delivery, computed
as the sum of orchestrator, Coding Agent (cumulative across all
attempts), DuckDB execution, and Interpretation Agent durations.
\textit{Token efficiency} is captured via total input
tokens ($\text{Tok}_{\text{in}}$) and output tokens
($\text{Tok}_{\text{out}}$) per query (summed across all agents), and the
per-query generation throughput (Thpt\,=\,output tokens
divided by LLM inference time for that query).
All values are reported as medians ($p_{50}$) over queries
pooled across the five runs.

\subsection{Evaluation Protocol}
\label{subsec:protocol}

All experiments are repeated over five independent runs per scenario and hardware setup after discarding one warm-up run to eliminate model-loading and caching effects. Quantitative metrics are averaged across runs, whereas qualitative evaluations follow the aggregation procedures described below.

\subsubsection{Quantitative Metrics}

All quantitative metrics are computed automatically for each of the 12 evaluation conditions (6 scenarios × 2 deployment tiers) and reported as the mean over the five runs.

\subsubsection{LLM-as-Judge}

One LLM evaluator performs three scoring passes per output,
yielding a median score per Likert dimension and a majority vote for binary
verdicts\footnote{DeepSeek-V4-Pro and Kimi-K2.6 were accessed via Ollama Cloud in May 2026.}:

\begin{itemize}
\item \textit{Interpretation quality}: It was evaluated by DeepSeek-V4-Pro for every valid interpretation ($\geq$10~characters, no pipeline
error), covering all 12~conditions.

\item \textit{Semantic Equivalence Review (SER):} SER candidates are
identified via majority-consistent filtering: a (query, condition) pair is
included only if it qualified as an SER candidate in $\geq$3 of the 5~runs, filtering out
stochastic one-off failures. A single representative run per eligible
pair is then reviewed. Evaluated also by DeepSeek-V4-Pro; the verdict is binary
(Yes/No) via majority vote of 3~passes.

\item \textit{Visualization quality}: It was evaluated by Kimi-K2.6 (vision-capable) for every successfully rendered visualization across all 12~conditions.
\end{itemize}

\subsubsection{Human Evaluation}

Four blinded domain experts, organized into two groups of two, evaluated the two ceiling configurations (\textsc{S+SC} and \textsc{M+Full}) across both deployment tiers under randomized system labels.

\begin{itemize}
\item \textit{Interpretation quality:} each group independently rates the
same subset of 28~intersection queries (those with valid interpretations
across all four conditions) under all four conditions, yielding
112~items per group (56~per evaluator).

\item \textit{SER:} majority-consistent filtering as above; all eligible
queries are included without intersection restriction (74~items total:
40~in group~A, 34~in group~B).

\item \textit{Visualization quality:} the intersection of queries with
renderable visualizations across all four conditions (35~items: 20~in
group~A, 15~in group~B).
\end{itemize}

\noindent For all human-evaluated dimensions, the best-performing run
(highest EX-R) per condition is selected as the representative one.

\section{Evaluation Results}
\label{sec:results_and_evaluation}

\subsection{Task Accuracy}
\label{subsec:task_accuracy}

  \newcommand{\tsd}[1]{{\tiny$_{\pm#1}$}}

  \begin{table*}[htbp]
      \centering
      \caption{Task accuracy across scenarios and tiers
          (mean\tsd{sd}, $N{=}5$ runs, 49 queries each).
          EX-S/P/R\,=\,Execution Strict/Partial/Relaxed (\%);
          EM\,=\,Exact Match (\%); CM\,=\,Component Matching $\in[0,1]$;
          ESR\,=\,Executable SQL Rate (\%); VER\,=\,Viz Execution Rate (\%).
          \textbf{Bold}\,=\,best per column.}
      \label{tab:task_accuracy_full}
      \setlength{\tabcolsep}{2pt}
      \renewcommand{\arraystretch}{1.2}
      \scriptsize
  \begin{tabular}{ll|ccc cc cc|ccc rc cc}
      \toprule
      & &
        \multicolumn{7}{c|}{\textbf{Tier 1} (Local, 9\,B)} &
        \multicolumn{7}{c}{\textbf{Tier 2} (Frontier, 397\,B)} \\
    \cmidrule(lr){3-9}\cmidrule(lr){10-16}
    & &
      \multicolumn{3}{c}{\textit{Execution}} &
      \multicolumn{2}{c}{\textit{Structural}} &
      \multicolumn{2}{c}{\textit{Runtime}} &
      \multicolumn{3}{c}{\textit{Execution}} &
      \multicolumn{2}{c}{\textit{Structural}} &
      \multicolumn{2}{c}{\textit{Runtime}} \\
    \cmidrule(lr){3-5}\cmidrule(lr){6-7}\cmidrule(lr){8-9}
    \cmidrule(lr){10-12}\cmidrule(lr){13-14}\cmidrule(lr){15-16}
    \textbf{Arch.} & \textbf{Scen.} &
      \rotatebox{60}{EX-S} & \rotatebox{60}{EX-P} & \rotatebox{60}{EX-R} &
      \rotatebox{60}{EM}   & \rotatebox{60}{CM}   &
      \rotatebox{60}{ESR}  & \rotatebox{60}{VER}  &
      \rotatebox{60}{EX-S} & \rotatebox{60}{EX-P} & \rotatebox{60}{EX-R} &
      \rotatebox{60}{EM}   & \rotatebox{60}{CM}   &
      \rotatebox{60}{ESR}  & \rotatebox{60}{VER}  \\
    \midrule
    \multirow{2}{*}{\textit{Single}}
      & S-Base
        & 24.5\tsd{3.2} & 25.3\tsd{2.3} & 32.7\tsd{4.1}
        & 3.7\tsd{1.7} & 0.273\tsd{0.018}
        & 53.5\tsd{3.0} & 27.6\tsd{4.2}
        & 36.7\tsd{1.4} & 37.1\tsd{2.2} & \textbf{54.7}\tsd{1.7}
        & \textbf{15.5}\tsd{2.3} & \textbf{0.422}\tsd{0.010}
        & 98.8\tsd{1.1} & 90.3\tsd{2.9} \\
      & S+SC
        & 26.5\tsd{4.8} & 28.2\tsd{4.7} & 36.3\tsd{7.8}
        & 10.6\tsd{1.7} & 0.309\tsd{0.020}
        & 68.6\tsd{16.8} & 49.1\tsd{7.1}
        & 35.5\tsd{4.2} & 35.5\tsd{4.2} & 53.1\tsd{4.8}
        & 12.7\tsd{2.7} & 0.403\tsd{0.021}
        & \textbf{99.6}\tsd{0.9} & 94.5\tsd{3.1} \\
    \midrule
    \multirow{4}{*}{\textit{Multi}}
      & M-Base
        & 28.2\tsd{2.7} & 29.8\tsd{2.3} & 35.5\tsd{2.3}
        & 15.1\tsd{1.8} & 0.335\tsd{0.019}
        & 57.6\tsd{3.9} & 28.3\tsd{9.9}
        & 34.7\tsd{3.8} & 35.5\tsd{3.7} & 50.6\tsd{3.7}
        & 11.0\tsd{1.8} & 0.362\tsd{0.045}
        & 95.1\tsd{4.0} & 88.3\tsd{5.2} \\
      & M+SC
        & 28.6\tsd{2.9} & 30.2\tsd{2.7} & 36.7\tsd{2.5}
        & \textbf{15.9}\tsd{3.9} & \textbf{0.359}\tsd{0.032}
        & \textbf{77.6}\tsd{5.2} & \textbf{56.6}\tsd{7.2}
        & 34.7\tsd{2.5} & 35.1\tsd{2.7} & 49.0\tsd{2.0}
        & 10.6\tsd{3.9} & 0.365\tsd{0.024}
        & \textbf{99.6}\tsd{0.9} & \textbf{96.6}\tsd{3.4} \\
      & M+KB
        & 29.4\tsd{3.4} & 31.4\tsd{4.0} & 38.4\tsd{4.4}
        & 9.8\tsd{2.2} & 0.337\tsd{0.020}
        & 58.8\tsd{3.4} & 31.0\tsd{6.5}
        & \textbf{38.0}\tsd{2.3} & \textbf{38.8}\tsd{2.9} & 54.3\tsd{4.2}
        & 10.6\tsd{3.4} & 0.365\tsd{0.031}
        & 94.7\tsd{2.7} & 84.8\tsd{7.2} \\
      & M+Full
        & \textbf{32.2}\tsd{5.3} & \textbf{33.9}\tsd{5.5} & \textbf{42.4}\tsd{3.9}
        & 12.2\tsd{2.0} & 0.357\tsd{0.030}
        & 74.7\tsd{2.7} & 47.6\tsd{4.5}
        & 35.1\tsd{4.7} & 35.9\tsd{5.3} & 53.1\tsd{5.2}
        & 9.0\tsd{1.8} & 0.349\tsd{0.024}
        & \textbf{99.6}\tsd{0.9} & 93.1\tsd{2.4} \\
    \bottomrule
    \end{tabular}
  \end{table*}

\noindent The two tiers exhibit qualitatively different patterns:

In Tier~1 (local, 9B), \textsc{M+Full} achieves the highest execution
accuracy across all three variants (EX-S~32.2\%, EX-P~33.9\%,
EX-R~42.4\%), a +6.1\,pp EX-R gain over the strongest single-agent
configuration (\textsc{S+SC}, 36.3\%). The highest ESR (77.6\%) and VER
(56.6\%), however, are obtained by \textsc{M+SC}, not by \textsc{M+Full}:
adding curated domain context on top of self-correction yields more
semantically accurate SQL but a small reduction in raw executability,
revealing a real trade-off in KB injection. Self-correction contributes a
large ESR improvement in both tracks (+15.1\,pp single, +20.0\,pp
multi), while domain knowledge adds +2.9\,pp EX-R independently of
self-correction.

  In Tier~2 (frontier, 397B), the pattern diverges. \textsc{M+KB} achieves
  the highest execution accuracy (EX-S~38.0\%, EX-P~38.8\%), overtaking
  \textsc{S-Base} (36.7\%) and indicating that domain knowledge injection
  remains beneficial even at frontier scale. However, the best EX-R is no
  longer \textsc{M+KB} (54.3\%) but \textsc{S-Base} (54.7\%); on this
  tier, the frontier model frequently returns valid but structurally
  divergent SQL that relaxed matching captures equally well without
  multi-agent orchestration. ESR is near-saturated ($\geq$94.7\%) across
  all multi-agent scenarios, leaving little room for the retry loop to act.
  \textsc{S-Base} retains the highest structural metrics (EM~15.5\%,
  CM~0.422), suggesting that the frontier model produces SQL closer to the
  gold structure when not guided by the orchestrator's plan and curated
  context. EM scores remain uniformly low ($\leq$15.9\%) across both tiers,
  confirming that the model rarely reproduces the exact gold SQL even when
  returning correct results.

  The $\Delta(\text{EX-R}\!-\!\text{EX-S})$ gap remains larger in
  Tier~2 (up to 18.0\,pp for \textsc{S-Base}) than in Tier~1 (up to
  10.2\,pp for \textsc{M+Full}): the frontier model frequently returns
  valid auxiliary columns that EX-Strict penalizes but that are
  operationally useful.

Table~\ref{tab:task_accuracy_full} reports the full results.

\subsubsection{Semantic Equivalence Rate}

\begin{table}[htbp]
\centering
\caption{Semantic Equivalence Rate (SER) per condition.
  LLM: majority vote, $N{=}3$ runs;
  HumEv: evaluator consensus (both raters agree).
  $n$\,=\,accepted\,/\,total candidates.}
\label{tab:ser-results}
\setlength{\tabcolsep}{4pt}
\footnotesize
\begin{tabular}{lll|rr|rr}
  \toprule
  & & &
    \multicolumn{2}{c|}{\textbf{Tier 1} (Local, 9\,B)} &
    \multicolumn{2}{c}{\textbf{Tier 2} (Frontier, 397\,B)} \\
\cmidrule(lr){4-5}\cmidrule(lr){6-7}
\textbf{Arch.} & \textbf{Scen.} & \textbf{Eval} &
  $n$ & Rate &
  $n$ & Rate \\
\midrule
\multirow{3}{*}{\textit{Single}}
  & S-Base & LLM   & 0/8   & 0.0\%  & 3/19  & 15.8\% \\
\cmidrule(l){2-7}
  & \multirow{2}{*}{S+SC}
           & LLM   & 1/20  & 5.0\%  & 7/24  & 29.2\% \\
  &        & HumEv & 6/20  & 30.0\% & 9/24  & 37.5\% \\
\midrule
\multirow{5}{*}{\textit{Multi}}
  & M-Base & LLM   & 3/9   & 33.3\% & 3/22  & 13.6\% \\
  & M+SC   & LLM   & 2/17  & 11.8\% & 4/24  & 16.7\% \\
  & M+KB   & LLM   & 1/10  & 10.0\% & 6/21  & 28.6\% \\
\cmidrule(l){2-7}
  & \multirow{2}{*}{M+Full}
           & LLM   & 1/16  & 6.2\%  & 7/23  & 30.4\% \\
  &        & HumEv & 4/16  & 25.0\% & 8/23  & 34.8\% \\
\bottomrule
\end{tabular}
\end{table}

Tier~2 yields higher LLM-judged SER than Tier~1 in most scenarios; within
Tier~2, \textsc{M+Full} (30.4\%), \textsc{S+SC} (29.2\%) and \textsc{M+KB}
(28.6\%) cluster closely. Tier~1 candidate pools are too small to support
a reliable ranking (e.g.\ \textsc{M-Base} at 33.3\% on only $n{=}9$
candidates). Human evaluators accept more alternatives than
the LLM judge: under strict consensus (both raters agree), \textsc{S+SC}
acceptance rises from 5.0\% to 30.0\% in Tier~1 and from 29.2\% to 37.5\%
in Tier~2; \textsc{M+Full} rises from 6.2\% to 25.0\% and from 30.4\% to
34.8\%, respectively. As an upper bound, single-rater acceptance reaches
70--100\%. This consistent gap between human and automated acceptance
indicates that automated evaluation systematically underestimates the
practical utility of generated SQL.

Table~\ref{tab:ser-results} details the per-condition breakdown.

\subsection{Self-Correction ROI}
\label{subsec:sc_roi}

The retry loop is highly effective when baseline
executability is low. In Tier~1, \textsc{S+SC} recovers $\Delta$ESR\,=\,22.4\,pp and $\Delta$EX-R\,=\,8.2\,pp; multi-agent scenarios show smaller but consistent gains ($\Delta$ESR 10.6-11.8\,pp, $\Delta$EX-R 2.0-3.3\,pp). In Tier~2, where baseline ESR already exceeds 95\%, the loop yields marginal returns ($\Delta$ESR\,$\leq$\,1.6\,pp, $\Delta$EX-R\,$\leq$\,0.4\,pp).

\subsection{Output Quality}
\label{subsec:output_quality}

  \begin{table*}[htbp]
    \centering
    \caption{Output quality scores (Likert 1--5). Grd\,=\,Groundedness,
        Rel\,=\,Relevance, Coh\,=\,Coherence, Aln\,=\,Viz-Text Alignment,
        Prs\,=\,Viz Presentation. LLM\,=\,LLM-as-a-Judge ($N{=}3$, median),
        HumEv\,=\,human evaluators (4 raters, 2 blinded groups;
        aggregate\,=\,mean of group means).
        Bold\,=\,best per metric within each tier and evaluation method.}
    \label{tab:output_quality}
    \setlength{\tabcolsep}{3pt}
    \renewcommand{\arraystretch}{1.05}
    \footnotesize
    \begin{tabular}{lll|ccc cc|ccc cc}
      \toprule
      & & &
        \multicolumn{5}{c|}{\textbf{Tier~1} (Local, 9\,B)} &
        \multicolumn{5}{c}{\textbf{Tier~2} (Frontier, 397\,B)} \\
      \cmidrule(lr){4-8}\cmidrule(lr){9-13}
      & & &
        \multicolumn{3}{c}{\textit{Interpretation}} &
        \multicolumn{2}{c|}{\textit{Visualization}} &
        \multicolumn{3}{c}{\textit{Interpretation}} &
        \multicolumn{2}{c}{\textit{Visualization}} \\
        \cmidrule(lr){4-6}\cmidrule(l){7-8}\cmidrule(lr){9-11}\cmidrule(l){12-13}
      \textbf{Arch.} & \textbf{Scen.} & \textbf{Eval} &
        Grd & Rel & Coh & Align & Pres &
        Grd & Rel & Coh & Align & Pres \\
      \midrule
      \multirow{3}{*}{\textit{Single}}
        & S-Base & LLM  & 3.53 & 4.45 & 4.56 & 3.30 & 3.02 & 4.03 & 4.65 & 4.86 & 3.87 & 3.41
  \\
      \cmidrule(l){2-13}
        & \multirow{2}{*}{S+SC}
                 & LLM   & 3.67 & 4.54 & 4.72 & 3.47 & 3.05 & 4.14 & 4.68 & 4.93 &
  \textbf{4.05} & 3.40 \\
        &        & HumEv & \textbf{4.41} & \textbf{3.80} & \textbf{4.30} & \textbf{4.15} &
  \textbf{3.27} & \textbf{4.57} & 3.91 & \textbf{4.14} & 3.27 &
    3.00 \\
      \midrule
      \multirow{5}{*}{\textit{Multi}}
        & M-Base & LLM  & \textbf{4.51} & \textbf{4.66} & \textbf{4.74} & 3.67 & 3.26 &
  \textbf{4.64} & \textbf{4.80} & 4.91 & 3.95 & 3.67 \\
        & M+SC   & LLM  & 4.46 & 4.61 & 4.69 & 3.69 & 3.26 & 4.59 & 4.79 & 4.90 & 3.95 & 3.71
  \\
        & M+KB   & LLM  & 4.21 & 4.63 & 4.72 & \textbf{3.92} & \textbf{3.56} & 4.29 & 4.68 &
  \textbf{4.94} & 3.93 & \textbf{3.75} \\
      \cmidrule(l){2-13}
        & \multirow{2}{*}{M+Full}
                 & LLM   & 4.13 & 4.51 & 4.52 & 3.57 & 3.35 & 4.24 & 4.74 & 4.91 & 3.97 & 3.70
   \\
        &        & HumEv & \textbf{4.41} & \textbf{3.80} & 3.86 & 3.69 & 3.19 & 4.54 &
  \textbf{4.11} & 4.05 & \textbf{4.29} &
    \textbf{3.17} \\
      \bottomrule
    \end{tabular}
  \end{table*}

In the LLM evaluation, multi-agent scenarios score consistently higher
\textit{Grd} than single-agent (\textsc{M-Base} 4.51/4.64 vs.\
\textsc{S-Base} 3.53/4.03), a gap that is conservative given that
S-Base receives the full uncurated KB whereas M-Base receives
none, supporting the agent-separation hypothesis.
Within the multi-agent track, however, adding curated KB context lowers \textit{Grd} in both tiers (\textsc{M-Base}
$\to$~\textsc{M+KB}: 4.51 $\to$ 4.21 in T1; 4.64 $\to$ 4.29 in T2):
KB-grounded contextual remarks added by the IA appear to be penalized
by the LLM judge as not literally traceable to the SQL result set.
\textit{Rel} and \textit{Coh} are uniformly high under LLM judgment
($\geq$4.45), with a Tier~2 ceiling near 4.9 for \textit{Coh}.
LLM-judged visualization scores are lower than interpretation scores
across all conditions (\textit{Align}~3.30-3.97, \textit{Pres}~3.02-3.75), with the
multi-agent advantage persisting
(\textsc{M-Base} Prs~3.26/3.67 vs.\ \textsc{S-Base} 3.02/3.41).

Human evaluators show a different pattern. \textit{Grd} is high and
comparable between \textsc{S+SC} (4.41/4.57) and \textsc{M+Full}
(4.41/4.54), narrowing the single-multi gap observed in the LLM
evaluation. Human-rated \textit{Rel} (3.80-4.11) and \textit{Coh}
(3.86-4.30) are consistently lower than LLM-judged values
(4.45-4.94), indicating a positivity bias in the LLM judge.
Visualization scores are moderate (\textit{Align}~3.27-4.29, \textit{Pres}~3.00-3.27) on
small per-condition samples ($n=6$-10), broadly consistent with the
LLM-judge range. 
Inter-rater reliability is low (weighted Cohen's $\kappa=0.04$--$0.11$ for Groundedness, $\leq 0.01$ for Relevance), reflecting the inherent subjectivity of Likert assessment on generated text
rather than a methodological flaw; we therefore report both group means and strict
consensus for binary metrics.

Table~\ref{tab:output_quality} presents the output quality scores.

\subsection{Response Time and Efficiency}
\label{subsec:response_time}

  \begin{table*}[t]
      \centering
        \caption{Response time and token efficiency across scenarios and tiers.
    $t_{\text{e2e}}$\,=\,end-to-end latency (s);
    Thpt\,=\,throughput (tok/s);
    Tok$_{\text{in}}$, Tok$_{\text{out}}$\,=\,input and output tokens per query.
    All values are medians ($p_{50}$) over pooled queries across 5~runs
    (245~queries per condition).
    Bold\,=\,best $t_{\text{e2e}}$ ($\downarrow$) and Thpt ($\uparrow$) per tier.}
      \label{tab:efficiency}
      \footnotesize{%
      \begin{tabular}{ll|rrrr|rrrr}
      \toprule
      & &
        \multicolumn{4}{c|}{\textbf{Tier 1} (Local, 9\,B)} &
        \multicolumn{4}{c}{\textbf{Tier 2} (Frontier, 397\,B)} \\
      \cmidrule(lr){3-6}\cmidrule(lr){7-10}
      \textbf{Arch.} & \textbf{Scen.} &
        $t_{\text{e2e}}$ & Thpt &
          Tok$_{\text{in}}$ & Tok$_{\text{out}}$ &
        $t_{\text{e2e}}$ & Thpt &
          Tok$_{\text{in}}$ & Tok$_{\text{out}}$ \\
      \midrule
      \multirow{2}{*}{\textit{Single}}
        & S-Base & \textbf{22.5} & 57.1 & 27\,796 & 1\,280 & \textbf{27.6} & 73.5 & 30\,315 &
  1\,965 \\
        & S+SC   & 31.9 & 60.2 & 34\,340 & 1\,914 & 28.2 & 72.4 & 30\,332 & 1\,926 \\
      \midrule
      \multirow{4}{*}{\textit{Multi}}
        & M-Base &  88.5 & 59.4 &  8\,425 & 5\,088 & 132.9 & 54.4 &  8\,911 & 6\,837 \\
        & M+SC   & 135.2 & \textbf{66.1} & 11\,953 & 8\,424 & 128.7 & 61.5 &  8\,956 & 7\,416
  \\
        & M+KB   & 144.4 & 46.3 & 49\,780 & 6\,494 & 474.0 & 20.7 & 49\,289 & 10\,208 \\
        & M+Full & 183.1 & 50.5 & 53\,047 & 9\,379 & 164.6 & \textbf{68.6} & 49\,592 & 10\,421
   \\
      \bottomrule
      \end{tabular}%
      }
    \end{table*}

Single-agent scenarios remain below 32\,s end-to-end in both tiers.
Multi-agent scenarios pay a coordination cost, \textsc{M+Full} reaches
183.1\,s (T1) and 164.6\,s (T2), driven primarily by KB injection:
\textsc{M+KB} and \textsc{M+Full} consume ${\sim}$49-53K input tokens
per query against ${\sim}$8--12K without KB, reducing Tier~1 throughput
from 59-66 to 46-50~tokens/s. The \textsc{M+KB} Tier~2 latency
(474.0\,s) is anomalously high relative to \textsc{M+Full} in the same
tier with near-identical input/output token volumes; we report it as an
outlier and refrain from causal interpretation. This value was not excluded from the reported mean; the median ($p_{50}$) latency metric used for all other comparisons is robust to such single-run anomalies.

Table~\ref{tab:efficiency} reports the latency and token efficiency metrics.

\section{Discussion}
\label{sec:discussion}

Full multi-agent orchestration shows its largest observed gains on resource-constrained local hardware (+6.1\,pp EX-R over the strongest single-agent baseline), where privacy also requires on-premises execution. On frontier models, individual components remain beneficial, but combining all of them no longer consistently outperforms simpler configurations, suggesting that orchestration should be applied selectively at larger model scales.

Self-correction and domain knowledge address complementary failure modes. Self-correction improves executability through iterative repair but saturates as baseline ESR increases, becoming marginal for the frontier model. Domain knowledge instead improves semantic correctness by guiding schema and threshold selection, remaining effective across both tiers. Their interaction is not fully additive: KB injection slightly reduces executability in some cases, indicating that additional context can occasionally interfere with schema linking. Finally, the frontier model often generates semantically correct queries containing auxiliary columns that conservative execution metrics penalize, while human evaluation confirms that multi-agent decomposition preserves explanation quality despite LLM-judge underestimation of groundedness.

Our evaluation is bounded by a single-deployment corpus drawn from one
industrial operator, a fixed zero-shot prompt configuration with no
few-shot examples provided to any agent, and a single LLM family
(Qwen~3.5); future work includes expanding the corpus across multiple
operators and geographies, cross-operator evaluation under different
regulatory frameworks, few-shot prompting with domain-specific
exemplars, agent fine-tuning on in-domain NL-to-SQL pairs, and
broadening the range of evaluated models to other open-weight families.

\section{Conclusion}

We present \textit{OpenMeter}, a privacy-preserving multi-agent system that enables non-technical utility personnel to query industrial smart meter databases in natural language using local open-weight models, ensuring that no telemetry or query content leaves the operator's infrastructure. A six-scenario ablation study on a real industrial partner's deployment shows that full multi-agent orchestration shows its largest observed benefits under resource-constrained local hardware, while at frontier scale individual components remain effective but selective activation is preferable.

Beyond system performance, our results suggest that automated evaluation metrics may underestimate the practical utility of LLM-generated SQL in industrial settings. The evaluation data are available upon request via Zenodo.

\section*{Acknowledgment}
This work has received funding from the European Union's Horizon Europe programme through the ENFIELD project (Grant Agreement No.~101120657). 
Know Center is a COMET competence center that is financed by
the Austrian Federal Ministry of Innovation, Mobility and Infrastruc-
ture (BMIMI), the Austrian Federal Ministry of Economy, Energy and
Tourism (BMWET), the Province of Styria, the Steirische Wirtschafts-
förderungsgesellschaft m.b.H. (SFG), the Vienna business agency and
the Standortagentur Tirol. The COMET programme is managed by the
Austrian Research Promotion Agency FFG. The authors thank Iskraemeco for providing the industrial smart meter datasets and domain expertise that enabled the evaluation of this work.

\end{document}